\documentclass[conference]{IEEEtran}
\IEEEoverridecommandlockouts
\usepackage{cite}
\usepackage{amsmath,amssymb,amsfonts}
\usepackage{algorithmic}
\usepackage{graphicx}
\usepackage{textcomp}
\usepackage{xcolor}
\def\BibTeX{{\rm B\kern-.05em{\sc i\kern-.025em b}\kern-.08em
    T\kern-.1667em\lower.7ex\hbox{E}\kern-.125emX}} 

\begin{document}

\title{\LARGE InsightSee: Advancing Multi-agent Vision-Language Models for Enhanced Visual Understanding}

\author{\IEEEauthorblockN{Huaxiang Zhang, Yaojia Mu, Guo-Niu Zhu*, Zhongxue Gan*}
\IEEEauthorblockA{\textit{Academy for Engineering and Technology, Fudan University, Shanghai 200433, China}\\
\textit{*Corresponding author: guoniu\_zhu@fudan.edu.cn, ganzhongxue@fudan.edu.cn}}}

\maketitle


\begin{abstract}
Accurate visual understanding is imperative for advancing autonomous systems and intelligent robots. Despite the powerful capabilities of vision-language models (VLMs) in processing complex visual scenes, precisely recognizing obscured or ambiguously presented visual elements remains challenging. To tackle such issues, this paper proposes InsightSee, a multi-agent framework to enhance VLMs' interpretative capabilities in handling complex visual understanding scenarios. The framework comprises a description agent, two reasoning agents, and a decision agent, which are integrated to refine the process of visual information interpretation. The design of these agents and the mechanisms by which they can be enhanced in visual information processing are presented. Experimental results demonstrate that the InsightSee framework not only boosts performance on specific visual tasks but also retains the original models' strength. The proposed framework outperforms state-of-the-art algorithms in 6 out of 9 benchmark tests, with a substantial advancement in multimodal understanding.
\end{abstract}


\section{Introduction}
Visual perception and understanding are crucial for robotics \cite{bao2023smart}. Various models have been proposed to enable robots to interpret their surroundings, make informed decisions, and interact with the world effectively \cite{mei2024gamevlm}. Among these models, the rise of large language models (LLMs) marks a significant advancement in natural language processing \cite{touvron2023llama}. It paves the way for the development of multimodal LLMs that combine linguistic and visual comprehension. For example, GPT-4V demonstrates impressive skills across a variety of tasks \cite{yang2023dawn}. However, these models often struggle with scenarios where objects are obscured or partially visible \cite{ha2022semantic}. They face challenges in identifying and interpreting intricate details within images.

To address such issues, recent advancements mainly focus on enhancing the specificity and accuracy of object recognition within images. Models such as Flamingo \cite{alayrac2022flamingo} incorporate the Perceiver Resampler to handle diverse visual inputs better. IDEFICS \cite{laurenccon2024obelics} improves feature extraction by integrating CLIP ViT features and herein enhances the model's understanding of complex visual scenes. Similarly, BLIP-2 \cite{li2023blip} introduces learnable queries with its Q-Former module to refine the interaction between visual and textual components. Qwen-VL \cite{bai2023qwen} compresses visual features into a fixed-length sequence by utilizing a cross-attention layer. 

The LLaMA-Adapter \cite{zhang2023llama} and LaVIN \cite{luo2024cheap} have pushed the boundaries further by integrating zero-gated self-attention layers and modality-specific adapters, respectively. The models maintained high fidelity in processing multimodal data. LLaVA-1.5 integrates advancements such as the use of a vision-language connector and extensive fine-tuning across various datasets to improve multimodal understanding \cite{liu2023improved}. 

However, current vision-language models (VLMs) perform poorly in recognizing obscured objects, particularly in cluttered or dynamic environments. Such limitations often lead to misinterpretations that may impair functionality and limit the deployment of autonomous systems \cite{song2024socially}. By contrast, human can typically recognize such obscured objects by relying on contextual cues and environmental information.

\begin{figure}[ht]
\centering
\includegraphics[width=8.5cm]{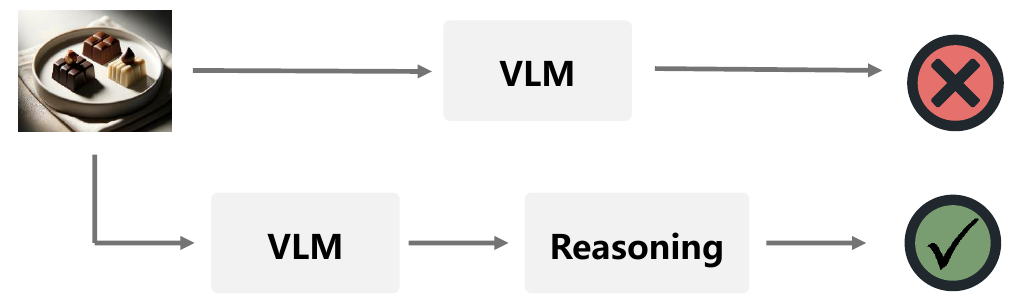}
\caption{Adding a reasoning process can enhance the VLM’s ability to resolve complex visual tasks.}
\label{figure_1}
\end{figure}

Inspired by the merit of the multi-agent strategy in group decision-making, this paper proposes InsightSee, a multi-agent framework to enhance VLMs' capabilities for interpreting complex images within visual scenes. As shown in Fig. \ref{figure_1}, adding a reasoning step after VLM can improve the model's ability to interpret complex images correctly. In this study, a description agent is introduced to analyze and describe images. Two reasoning agents are developed to interpret and synthesize the output of the description agent. An adversarial process is presented to refine these interpretations. A decision agent is suggested to evaluate the analysis and formulate a conclusion.

\begin{figure*}[ht]
\centering
\includegraphics[width=18.5cm]{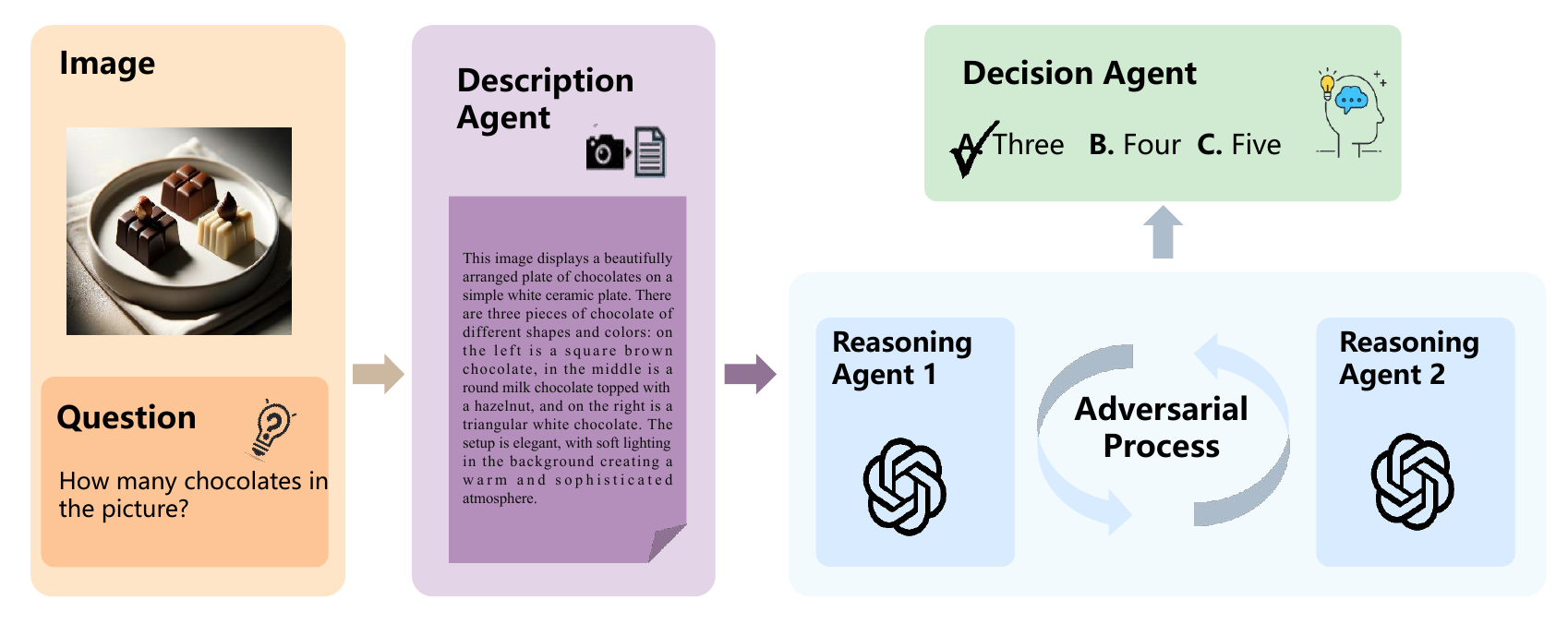}
\caption{Overview of the InsightSee} 
\label{figure_2}
\end{figure*}
The main contributions of this paper are summarized as follows.

(1) This study proposes InsightSee, a multi-agent framework for visual understanding and interpreting. Multi-agent strategies are introduced to enhance VLMs in intricate tasks.

(2) An adversarial reasoning mechanism is presented to evaluate and refine the interpretations from multiple agents. The results are optimized through such an adversarial process.

(3) Multiple experiments on public datasets are conducted to evaluate the efficacy of the proposed method. Experimental results demonstrate the superiority of the proposed InsightSee, with a higher accuracy on benchmark tasks.

\section{Methodology}
This section presents InsightSee, a multi-agent framework to address the challenges faced in complex image recognition and understanding. As illustrated in Fig. \ref{figure_2}, the framework comprises a description agent, two reasoning agents, and a decision agent. Each agent refers to a VLM. The design and functionality of each component are detailed as follows.

\subsection{Framework of the Proposed InsightSee} 
The InsightSee framework is presented to handle complex image analysis tasks through structured collaboration between the description and reasoning agents. The decision agent is then used to facilitate the synthesis and decision-making process. This structured interaction provides a comprehensive
way for image interpretation, which is especially valuable in challenging scenarios involving obscured or partially visible objects. The process is conducted as follows.

The InsightSee begins with the description agent, which receives the input image and associated queries. It leverages VLMs to conduct an in-depth image analysis and identify key visual elements. The description agent provides descriptions from both a comprehensive and detailed perspective. In terms of details, the agent thoroughly describes the various aspects of the regions of interest, such as size, shape, and color. From a comprehensive viewpoint, it outlines the main content of the scene to capture essential elements that contribute to an overall understanding.

Following the description agent, two reasoning agents delve into a more complex analysis. Using the output from the description agent, along with the image and related queries as inputs, the reasoning agents engage in adversarial reasoning.  They develop and refine their interpretations by iteratively challenging each other's views. The process is concluded either upon reaching a consensus or after a predetermined number of rounds.

After the adversarial process, the decision agent is adopted to synthesize and consolidate the analyses from the reasoning agents. It collects and coordinates extensive inputs to enhance the accuracy of the decision. The decision agent formulates a conclusion that captures the reasoned insights without delving into the specific analytical process used by the reasoning agents.

\subsection{Description Agent}
The description agent is put forward to enhance the interpretation of images by leveraging VLMs for a thorough analysis. This agent systematically dissects and describes images from both global and detailed perspectives, which adopts a chain-of-thought \cite{wei2022chain} strategy to enhance the depth and accuracy of its analysis. It utilizes specific prompting techniques to achieve a nuanced understanding of visual data.

\textbf{Global Perspective:}
The initial step involves identifying the main objects and characters within the image. Then, the agent examines interactions and attributes among these elements to infer their social and cultural properties. Lastly, it integrates such information with the scene's background to outline the societal or cultural messages conveyed.

\textbf{Detailed Perspective:}
For a detailed analysis, the description agent begins with a comprehensive observation of the scene to establish context. It then identifies key regions or elements that are critical to the query at hand. Focusing on these areas, it examines details such as shape, color, and distinct markers.

\subsection{Reasoning Agent} 
The reasoning agent is introduced to interpret and synthesize the detailed descriptions provided by the description agent to enhance the understanding of complex images. It leverages object attributes and environmental context to improve reasoning accuracy, particularly in scenarios with obscured or ambiguously presented objects. The prompts for the reasoning agent are divided into two parts: analysis and adversarial reasoning mechanism.

\begin{figure*}[ht]
\centering
\includegraphics[width=17.5cm]{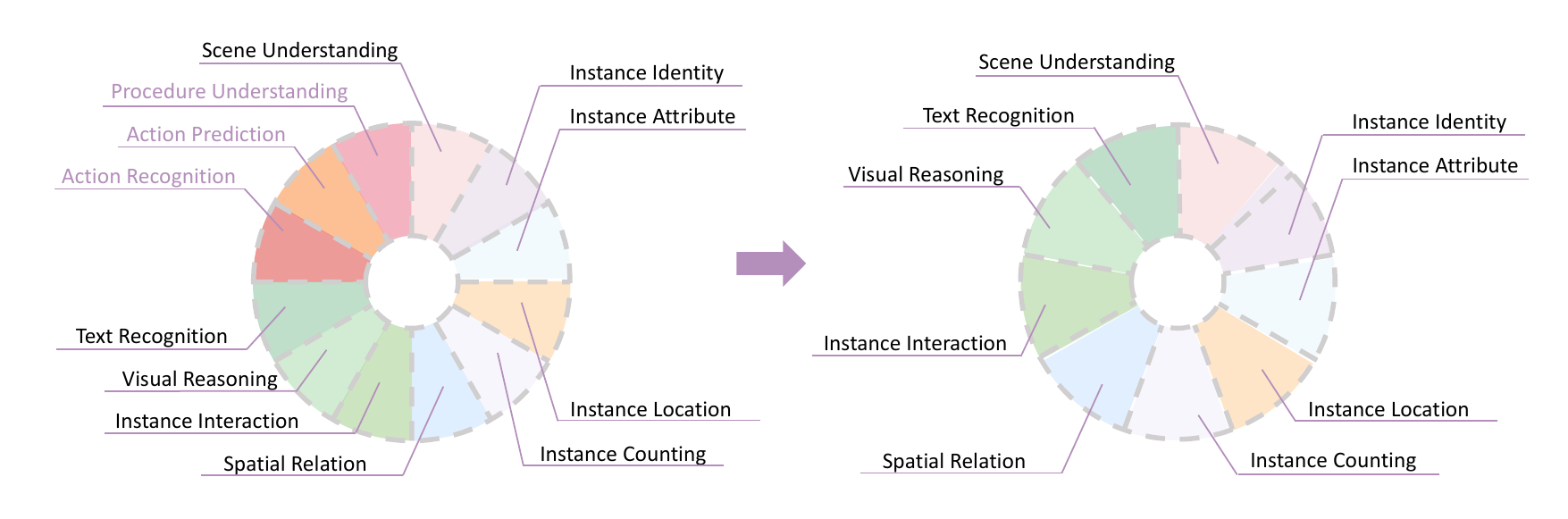}
\caption{Distribution of the dataset}
\label{figure_3}
\end{figure*}

\textbf{Analysis:}
 The agent analyzes the image by focusing on specific object attributes such as size, shape, color, and positioning, as well as the broader environmental context including lighting, spatial arrangement, and interactions between elements. This comprehensive analysis helps the agent formulate hypotheses about the scene based on detailed observations and overall scene interpretation. These functions are guided by specific prompts designed to ensure thorough analysis.

\textbf{Adversarial Reasoning Mechanism:} 
The reasoning agent adopts an adversarial reasoning mechanism to enhance their analyses. In this process, each agent is asked to provide interpretations and predictions about the scene independently, which are then evaluated and refined through structured debate or comparison. This adversarial interaction prompts each agent to critically assess its hypotheses and adjust its reasoning based on the strengths and weaknesses revealed during the debate.

\subsection{Decision Agent}
The decision agent is proposed to synthesize the analyses provided by the reasoning agents and make the final decision. It consolidates the various interpretations from the reasoning agents to formulate a conclusion about the visual understanding tasks. When the reasoning agents reach conflicting conclusions, the decision agent performs a voting mechanism to resolve these discrepancies. Each agent's conclusion is considered a vote. The final decision is determined by the majority vote. 

\section{Experiments}
This section provides the details of the experiments. The datasets, baseline methods, and implementation details are outlined.

\subsection{Datasets}
As shown in Fig. \ref{figure_3}, a modified version of the SEED-Bench dataset \cite{li2023seed} is used in this study. This dataset is initially proposed to evaluate the understanding and reasoning capabilities of multimodal LLMs across various visual and textual tasks. SEED-Bench consists of a wide range of multiple-choice questions distributed across twelve dimensions. This study mainly focuses on spatial image comprehension. Accordingly, we select nine dimensions: scene understanding, instance identity, instance attributes, instance location, instance counting, spatial relation, instance interaction, visual reasoning, and text recognition. We exclude dimensions related to temporal understanding, such as procedure understanding, action prediction, and action recognition, to concentrate on spatial reasoning capabilities.

\begin{figure*}[ht]
\centering
\includegraphics[width=17.5cm]{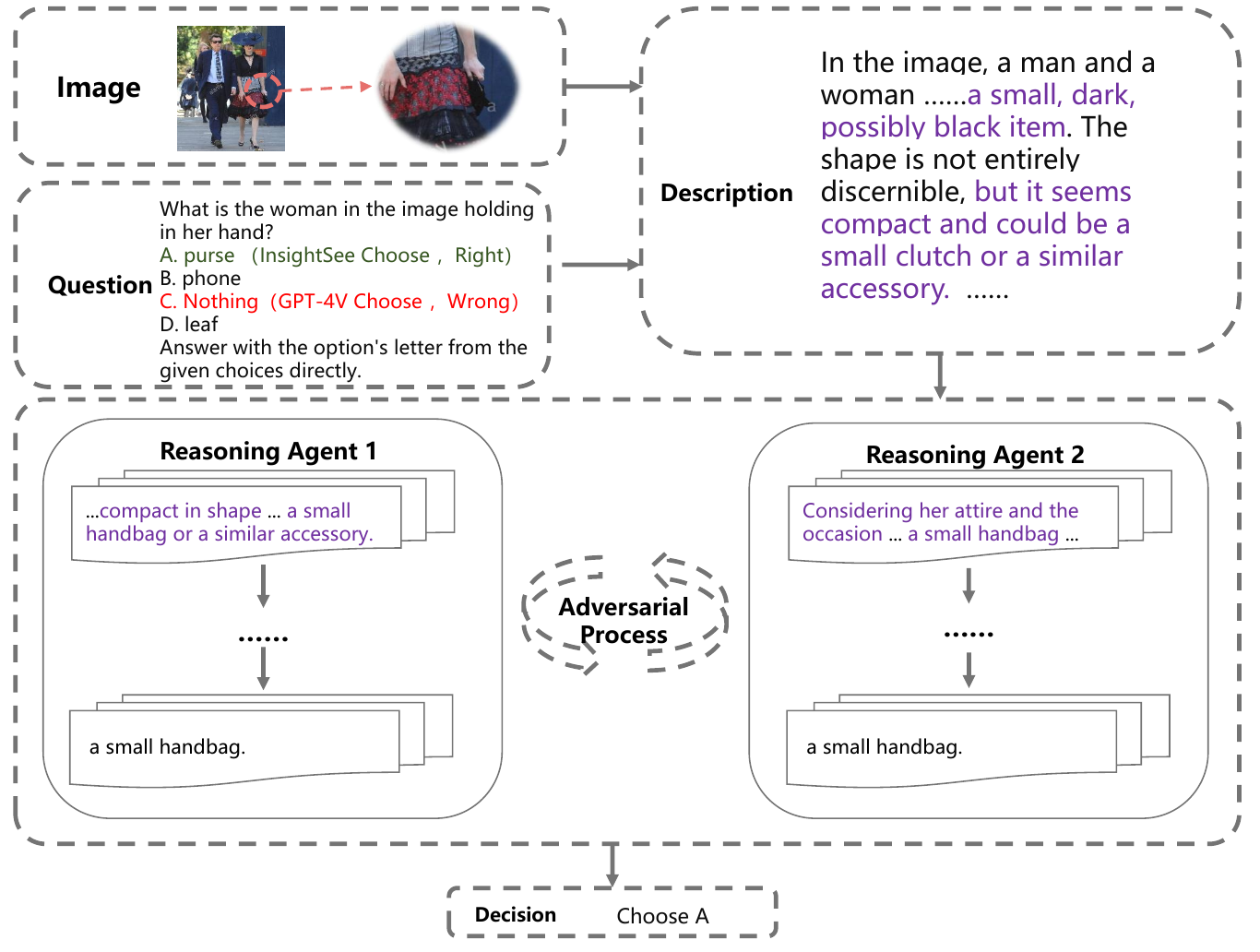}
\caption{Real experimental scenarios}
\label{figure_4}
\end{figure*}

\begin{table*}[htbp]
\caption{Experimental results \cite{seedbench2023}}
\begin{center}
\begin{tabular}{|c|c|c|c|c|c|c|c|c|c|c|}
\hline
\textbf{Model} & \textbf{SU} & \textbf{IIden} & \textbf{IA} & \textbf{IL} & \textbf{ICount} & \textbf{SR} & \textbf{IInter} & \textbf{ViR} & \textbf{TR} & \textbf{Average} \\
\hline
InstructBLIP-Vicuna\cite{dai2024instructblip}& 60.2\% & 58.9\% & 65.6\% & 43.6\% & 57.2\% & 40.3\% & 52.6\% & 47.7\% & 43.5\% & 52.14\% \\
\hline
InstructBLIP\cite{dai2024instructblip} & 60.3\% & 58.5\% & 63.4\% & 40.6\% & 58.4\% & 38.7\% & 51.6\% & 45.9\% & 25.9\% & 49.26\% \\
\hline
Qwen-VL\cite{bai2023qwen}& 71.2\% & 66.4\% & 67.7\% & 53.5\% & 44.8\% & 43.8\% & 62.9\% & 74.9\% & 51.2\% & 59.60\% \\
\hline
LLaVA-1.5\cite{liu2023improved} & 74.9\% & 71.3\% & 68.9\% & 63.5\% & 61.3\% & 51.4\% & 73.2\% & 77.0\% & \textbf{60.5\%} & 66.89\% \\
\hline
ShareGPT4V-13B\cite{chen2023sharegpt4v}& 75.9\% & 74.1\% & 73.5\% & 66.8\% & 62.4\% & 54.8\% & 75.3\% & 77.3\% & 46.5\% & 67.40\% \\
\hline
InternVL-Chat-V1.2-Plus\cite{chen2023internvl} & 80.2\% & \textbf{80.0\%} & 77.8\% & \textbf{71.3\%} & \textbf{72.3\%} & 63.3\% & 77.3\% & 79.8\% & 50.0\% & 72.44\% \\
\hline
GPT-4V\cite{openai2023gpt4v} & 77.5\% & 73.9\% & 70.6\% & 61.8\% & 56.8\% & 56.9\% & 74.2\% & 78.5\% & 57.6\% & 67.53\% \\
\hline
InsightSee-GPT4V(Ours) & \textbf{82.1\%} & \textbf{80.0\%} & \textbf{79.3\%} & 70.7\% & 68.6\% & \textbf{63.6\%} & \textbf{80.6\%} & \textbf{87.7\%} & 57.6\% & \textbf{74.47\%} \\
\hline
\end{tabular}
\end{center}
\label{table1}
\end{table*}

\begin{figure*}[ht]
\centering
\includegraphics[width=17.5cm]{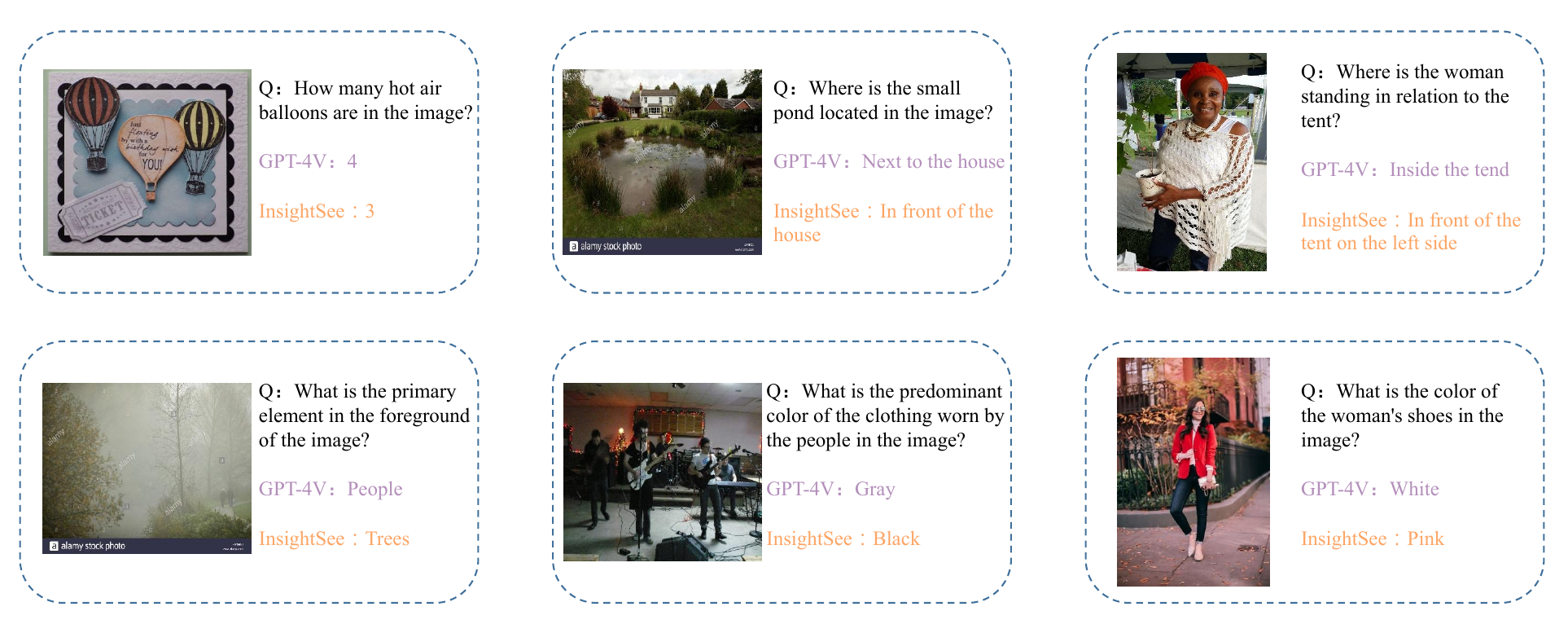}
\caption{Examples of which GPT-4V fails while InsightSee succeeds.}
\label{figure_5}
\end{figure*}

The SEED-Bench dataset distributes unevenly across different dimensions, with categories ranging from 97 questions in instance interaction to 3158 questions in scene understanding. This variation could distort performance metrics, which leads to a misleading assessment of their capabilities in visual scene understanding. To address such issues, this study uses a task-average approach, which averages the model’s performance across the selected nine dimensions before calculating the overall average. The task-average tactics reduce bias from question distribution disparities and obtain a clearer view of the model's spatial reasoning abilities. Additionally, SEED-Bench operates without human or GPT intervention during evaluations, which can enhance the efficiency and objectivity of the process.

\subsection{Baseline Methods} To evaluate the effectiveness of the proposed InsightSee framework, comparative studies are conducted on the experimental dataset. Some state-of-the-art algorithms include InstructBLIP-Vicuna \cite{dai2024instructblip}, InstructBLIP \cite{dai2024instructblip}, Qwen-VL \cite{bai2023qwen}, LLaVA-1.5 \cite{liu2023improved}, GPT-4V \cite{openai2023gpt4v}, InternVL-Chat-V1.2-Plus \cite{chen2023internvl}, and ShareGPT-4V-13B \cite{chen2023sharegpt4v} are introduced as the baseline methods.

\subsection{Implementation Details} 
First of all, the description agent is implemented to initiate the analysis by conducting a detailed scan of the visual input. Based on the chain-of-thought strategy, the description agent deconstructs the image, gets a broad overview of the scene, and gradually narrows it down to specific details. The prompt guides the agent in identifying and inferring obscured objects by examining visible elements and integrating these observations with environmental cues. 

After that, the reasoning agents operate on the output provided by the description agent and conduct their independent analysis. These reasoning agents are asked to engage in an adversarial interaction process, where they iterative challenge each other's interpretations and hypotheses. The adversarial process runs several rounds. In each round, one reasoning agent presents a proposal, which the other agents evaluate and challenge. The process ends after a preset number of rounds or once a consensus is reached.

Finally, the decision agent finalizes the decision-making process by synthesizing insights from the VLMs and the analyses provided by the reasoning agents. It uses meticulous prompts to guide the reasoning process. If the conclusions provided by the reasoning agents are contradictory even after adversarial interaction, conflict resolution rules will be introduced to finalize a final solution. The decision agent integrates the image descriptions and its own analysis as a neutral third-party input to adjudicate and arrive at the final decision. 

Furthermore, a thousand image understanding questions are selected from the dataset and used for testing. Each question is tested three times to alleviate the randomness in the content generated by agents. A question is considered correctly answered when at least two out of the three responses are correct. Accuracy is introduced as the evaluation metric, which is defined as the percentage of the questions correctly answered out of the total.

\section{Results}
\subsection{Experimental Results}
In the experiment, GPT-4V is selected as the VLMs of the agents in the proposed InsightSee. Table \ref{table1} presents the models' experimental results on the nine dimensions of the SEED-Bench dataset. As shown in Table \ref{table1}, the proposed InsightSee achieves an accuracy of 82.1\% on scene understanding, 80\% on instance identity, 79.3\% on instance attributes, 70.7\% on instance location, 68.6\% on instance counting, 63.6\% on spatial relation, 80.6\% on instance interaction, 87.7\% on visual reasoning, and 57.6\% on text recognition. The proposed framework outperforms the baseline methods on most of the evaluation dimensions. It gets the highest average accuracy and achieves the best in six evaluation dimensions.

From the experimental results, it can be observed that the proposed InsightSee framework significantly enhances VLMs' performance in multiple evaluation dimensions. Utilizing GPT-4V as the base VLM, the InsightSee framework achieves considerable improvements in the dimensions of instance attributes, instance location, instance counting, and visual reasoning, with accuracy gains of around 9\% or more. Thanks to the structured collaboration between the description agent and the two reasoning agents, the multi-agent framework outperforms the baseline methods with substantial enhancements.

Despite its generally strong performance, InsightSee shows limitations, especially in text recognition. The InsightSee and GPT-4V achieve the same accuracy in this dimension. As the InsightSee framework is proposed for parsing and interpreting complex visual and spatial data rather than detailed textual content, its architecture prioritizes visual elements over textual ones, which leads to less accuracy in text recognition.

\subsection{Case Study}
Fig. \ref{figure_4} demonstrates how InsightSee enhances the image understanding capabilities of GPT-4V.  Through the multi-agent framework, the proposed InsightSee is able to analyze and interpret complex visual scenes more deeply. Here, we illustrate the advantages and processes of our framework through a specific case study. As illustrated in Fig. \ref{figure_4}, we analyze a case study where the input is an image depicting a woman whose right hand is holding a purse that is significantly obscured. The question posed to the model inquires what the woman is holding in her hand.

\textbf{Initial Processing by the Description Agent:} The description agent begins by analyzing the visual input to identify critical elements despite occlusions. In this case, it recognizes a compact object in the woman’s hand, which could not be fully discerned. The agent describes it as a small and dark item, which is possibly a clutch or a similar accessory. Although the shape is obscured, sufficient features are visible to hypothesize the category of the object.

\textbf{Enhanced Analysis Through Reasoning Agents:} Two reasoning agents analyze the information provided by the description agent. The first reasoning agent focuses on the object's shape and compactness and tries to narrow it down to a small handbag or a similar accessory based on visible contours and the proportions in relation to the woman's hand. The second reasoning agent considers additional context, such as the woman's attire and the likely social setting, to find the cues to support the hypothesis posed by the first reasoning agent.

\textbf{Adversarial Process and Decision Making:} The two reasoning agents engage in an adversarial interaction to challenge and refine their hypotheses. This process fosters a dynamic debate between the agents and prompts them to focus on the most probable interpretation based on combined visual and contextual analysis.

\textbf{Decision Synthesis:} The decision agent synthesizes the inputs from both reasoning agents. Given that their analyses converge on identifying the object as a small handbag, the decision agent is supposed to validate this outcome as the most probable answer to the question posed.

Fig. \ref{figure_5} shows examples where GPT-4V fails but InsightSee succeeds. These examples involve hidden objects, complex spatial relations, and subtle visual details. The success of the InsightSee can be explained as follows.

\textbf{Teamwork:} The description and reasoning agents takes account for different parts of the image, respectively. They work as a team, which enable the framework a thorough analysis of the image.

\textbf{Adversarial Reasoning:} The agents challenge each other's ideas, which refines their conclusions through multiple rounds of debate.

\textbf{Combining Details and Context:} The agents look at the small details and the overall context, which helps clear up ambiguities.

With the aid of the above reasons, the proposed InsightSee performs better than GPT-4V on complex visual tasks.

\section{Conclusion}
This paper proposed InsighSee, a multi-agent framework to enhance the capabilities of VLMs in complex visual understanding tasks. In this framework, GPT-4V was used as the base model of the agent. A description agent was introduced to describe and analyze images. Two reasoning agents were put forward to interpret and synthesize the outputs of the description agent. A decision agent was suggested to evaluate the results and formulate a conclusion. Experimental results on public datasets demonstrated the superiority of the proposed InsightSee, with a higher accuracy over baseline methods.

Although the proposed InsightSee works well on most tasks, it performs poorly on text recognition. In the future, we will optimize the framework by integrating sophisticated optical character recognition tools to enhance its performance on such tasks.

\section*{Acknowledgement}
This work is supported by Shanghai Municipal Science and Technology Major Project under Grant 2021SHZDZX0103 and Key Project of Comprehensive Prosperity Plan of Fudan University under Grant XM06231744.

\bibliographystyle{IEEEtran}
\bibliography{ref}

\begin{thebibliography}{10}
\providecommand{\url}[1]{#1}
\csname url@samestyle\endcsname
\providecommand{\newblock}{\relax}
\providecommand{\bibinfo}[2]{#2}
\providecommand{\BIBentrySTDinterwordspacing}{\spaceskip=0pt\relax}
\providecommand{\BIBentryALTinterwordstretchfactor}{4}
\providecommand{\BIBentryALTinterwordspacing}{\spaceskip=\fontdimen2\font plus
\BIBentryALTinterwordstretchfactor\fontdimen3\font minus \fontdimen4\font\relax}
\providecommand{\BIBforeignlanguage}[2]{{%
\expandafter\ifx\csname l@#1\endcsname\relax
\typeout{** WARNING: IEEEtran.bst: No hyphenation pattern has been}%
\typeout{** loaded for the language `#1'. Using the pattern for}%
\typeout{** the default language instead.}%
\else
\language=\csname l@#1\endcsname
\fi
#2}}
\providecommand{\BIBdecl}{\relax}
\BIBdecl

\bibitem{bao2023smart}
Z.~Bao, G.-N. Zhu, W.~Ding, Y.~Guan, W.~Bai, and Z.~Gan, ``A smart interactive camera robot based on large language models,'' in \emph{2023 IEEE International Conference on Robotics and Biomimetics (ROBIO)}.\hskip 1em plus 0.5em minus 0.4em\relax IEEE, 2023, pp. 1--6.

\bibitem{mei2024gamevlm}
A.~Mei, J.~Wang, G.-N. Zhu, and Z.~Gan, ``{GameVLM}: A decision-making framework for robotic task planning based on visual language models and zero-sum games,'' \emph{arXiv preprint arXiv:2405.13751}, 2024.

\bibitem{touvron2023llama}
H.~Touvron, T.~Lavril, G.~Izacard, X.~Martinet, M.-A. Lachaux, T.~Lacroix, B.~Rozi{\`e}re, N.~Goyal, E.~Hambro, F.~Azhar \emph{et~al.}, ``{LLaMA}: Open and efficient foundation language models,'' \emph{arXiv preprint arXiv:2302.13971}, 2023.

\bibitem{yang2023dawn}
Z.~Yang, L.~Li, K.~Lin, J.~Wang, C.-C. Lin, Z.~Liu, and L.~Wang, ``The dawn of {LMMs}: Preliminary explorations with {GPT-4V(ision)},'' \emph{arXiv preprint arXiv:2309.17421}, vol.~9, no.~1, p.~1, 2023.

\bibitem{ha2022semantic}
H.~Ha and S.~Song, ``Semantic abstraction: Open-world {3D} scene understanding from {2D} vision-language models,'' \emph{arXiv preprint arXiv:2207.11514}, 2022.

\bibitem{alayrac2022flamingo}
J.-B. Alayrac, J.~Donahue, P.~Luc, A.~Miech, I.~Barr, Y.~Hasson, K.~Lenc, A.~Mensch, K.~Millican, M.~Reynolds \emph{et~al.}, ``Flamingo: a visual language model for few-shot learning,'' \emph{Advances in Neural Information Processing Systems}, vol.~35, pp. 23\,716--23\,736, 2022.

\bibitem{laurenccon2024obelics}
H.~Lauren{\c{c}}on, L.~Saulnier, L.~Tronchon, S.~Bekman, A.~Singh, A.~Lozhkov, T.~Wang, S.~Karamcheti, A.~Rush, D.~Kiela \emph{et~al.}, ``Obelics: An open web-scale filtered dataset of interleaved image-text documents,'' \emph{Advances in Neural Information Processing Systems}, vol.~36, 2024.

\bibitem{li2023blip}
J.~Li, D.~Li, S.~Savarese, and S.~Hoi, ``{BLIP-2}: Bootstrapping language-image pre-training with frozen image encoders and large language models,'' in \emph{International Conference on Machine Learning}.\hskip 1em plus 0.5em minus 0.4em\relax PMLR, 2023, pp. 19\,730--19\,742.

\bibitem{bai2023qwen}
J.~Bai, S.~Bai, S.~Yang, S.~Wang, S.~Tan, P.~Wang, J.~Lin, C.~Zhou, and J.~Zhou, ``{Qwen-VL}: A frontier large vision-language model with versatile abilities,'' \emph{arXiv preprint arXiv:2308.12966}, 2023.

\bibitem{zhang2023llama}
R.~Zhang, J.~Han, C.~Liu, P.~Gao, A.~Zhou, X.~Hu, S.~Yan, P.~Lu, H.~Li, and Y.~Qiao, ``{LLaMA-Adapter}: Efficient fine-tuning of language models with zero-init attention,'' \emph{arXiv preprint arXiv:2303.16199}, 2023.

\bibitem{luo2024cheap}
G.~Luo, Y.~Zhou, T.~Ren, S.~Chen, X.~Sun, and R.~Ji, ``Cheap and quick: Efficient vision-language instruction tuning for large language models,'' \emph{Advances in Neural Information Processing Systems}, vol.~36, 2024.

\bibitem{liu2023improved}
H.~Liu, C.~Li, Y.~Li, and Y.~J. Lee, ``Improved baselines with visual instruction tuning,'' \emph{arXiv preprint arXiv:2310.03744}, 2023.

\bibitem{song2024socially}
D.~Song, J.~Liang, A.~Payandeh, X.~Xiao, and D.~Manocha, ``Socially aware robot navigation through scoring using vision-language models,'' \emph{arXiv preprint arXiv:2404.00210}, 2024.

\bibitem{wei2022chain}
J.~Wei, X.~Wang, D.~Schuurmans, M.~Bosma, F.~Xia, E.~Chi, Q.~V. Le, D.~Zhou \emph{et~al.}, ``Chain-of-thought prompting elicits reasoning in large language models,'' \emph{Advances in Neural Information Processing Systems}, vol.~35, pp. 24\,824--24\,837, 2022.

\bibitem{li2023seed}
B.~Li, R.~Wang, G.~Wang, Y.~Ge, Y.~Ge, and Y.~Shan, ``{SEED-Bench}: Benchmarking multimodal {LLMs} with generative comprehension,'' \emph{arXiv preprint arXiv:2307.16125}, 2023.

\bibitem{seedbench2023}
{AILab-CVC}, ``{SEED-Bench} leaderboard,'' \url{https://huggingface.co/spaces/ AILab-CVC/SEED-Bench\_Leaderboard}, 2023, accessible Date: 06/05/ 2024.

\bibitem{dai2024instructblip}
W.~Dai, J.~Li, D.~Li, A.~M.~H. Tiong, J.~Zhao, W.~Wang, B.~Li, P.~N. Fung, and S.~Hoi, ``{InstructBLIP}: Towards general-purpose vision-language models with instruction tuning,'' \emph{Advances in Neural Information Processing Systems}, vol.~36, 2024.

\bibitem{chen2023sharegpt4v}
L.~Chen, J.~Li, X.~Dong, P.~Zhang, C.~He, J.~Wang, F.~Zhao, and D.~Lin, ``{ShareGPT4V}: Improving large multi-modal models with better captions,'' \emph{arXiv preprint arXiv:2311.12793}, 2023.

\bibitem{chen2023internvl}
Z.~Chen, J.~Wu, W.~Wang, W.~Su, G.~Chen, S.~Xing, Z.~Muyan, Q.~Zhang, X.~Zhu, L.~Lu \emph{et~al.}, ``{InternVL}: Scaling up vision foundation models and aligning for generic visual-linguistic tasks,'' \emph{arXiv preprint arXiv:2312.14238}, 2023.

\bibitem{openai2023gpt4v}
{OpenAI}, ``{GPT-4V (ision) System Card},'' \url{https://cdn.openai.com/papers/ GPTV\_System\_Card.pdf}, 2023.

\end{thebibliography}
\end{document}